\documentclass[letterpaper, 10 pt, conference]{ieeeconf}  % Comment this line out
\IEEEoverridecommandlockouts                              % This command is only
\usepackage{amsmath,amssymb,amsfonts, mathtools}
\usepackage{graphicx}
\usepackage{bm}
\usepackage{epstopdf}
\usepackage{todonotes}
\usepackage{placeins}
\usepackage{bm}
\usepackage{cite}
\let\labelindent\relax
\usepackage[inline]{enumitem}
\usepackage{multicol}
\usepackage{tikz}
\usepackage{caption}
\usepackage{subcaption}
\usepackage{soul}
\usepackage{wrapfig}
\usepackage{tabto}
\usepackage{comment}
\usepackage{xcolor}
\usepackage[ruled,vlined]{algorithm2e}
\usepackage{hyperref}
\usepackage{booktabs} % for tables

\usepackage{enumitem}

\newcommand{\vect}[1]{\boldsymbol{#1}}

\DeclarePairedDelimiter\abs{\lvert}{\rvert}%
\DeclarePairedDelimiter\norm{\lVert}{\rVert}%

\makeatletter
\let\oldabs\abs
\def\abs{\@ifstar{\oldabs}{\oldabs*}}
\let\oldnorm\norm
\def\norm{\@ifstar{\oldnorm}{\oldnorm*}}
\makeatother

\title{\LARGE \bf Similarity-Aware Skill Reproduction based on \\ Multi-Representational Learning from Demonstration}

\author{Brendan Hertel and S. Reza Ahmadzadeh
	\thanks{Persistent Autonomy and Robot Learning (PeARL) Lab, University of Massachusetts Lowell, Lowell, MA, 01854. Email: {\tt\small brendan\_hertel@student.uml.edu,reza@cs.uml.edu}}%
}
	
\begin{document}
\bstctlcite{IEEEexample:BSTcontrol}

\maketitle
\thispagestyle{empty}
\pagestyle{empty}

%%%%%%%%%%%%%%%%%%%%%%%%%%%%%%%%%%%%%%%%%%%%%%%%%%%%%%%%%%%%%%%%%%%%%%%%%%%%%%%%
\begin{abstract}

Learning from Demonstration (LfD) algorithms enable humans to teach new skills to robots through demonstrations. The learned skills can be robustly reproduced from the identical or near boundary conditions (e.g., initial point). However, when generalizing a learned skill over boundary conditions with higher variance, the similarity of the reproductions changes from one boundary condition to another, and a single LfD representation cannot preserve a consistent similarity across a generalization region. We propose a novel similarity-aware framework including multiple LfD representations and a similarity metric that can improve skill generalization by finding reproductions with the highest similarity values for a given boundary condition. Given a demonstration of the skill, our framework constructs a similarity region around a point of interest (e.g., initial point) by evaluating individual LfD representations using the similarity metric. Any point within this volume corresponds to a representation that reproduces the skill with the greatest similarity. We validate our multi-representational framework in three simulated and four sets of real-world experiments using a physical 6-DOF robot. We also evaluate 11 different similarity metrics and categorize them according to their biases in 286 simulated experiments.

\end{abstract}

%%%%%%%%%%%%%%%%%%%%%%%%%%%%%%%%%%%%%%%%%%%%%%%%%%%%%%%%%%%%%%%%%%%%%%%%%%%%%%%%
\section{Introduction}
\label{sec:intro}

As robots become increasingly present in the world, they are expected to be able to learn, reproduce, and generalize new skills effortlessly and robustly. Learning from Demonstration (LfD)~\cite{ravichandar2020recent} is one of the most intuitive robot learning methods that enables humans to teach new primitive skills or movements (e.g., reaching) to robots only by providing examples, bypassing the need for manual programming of movements into the system. Several LfD algorithms with various representations ranging from statistical~\cite{hertel2021TLFSD} and probabilistic models~\cite{ProMP} to geometric objects~\cite{TLGC} and dynamical systems~\cite{pastorDMP2009} have been developed and evaluated~\cite{ravichandar2020recent}. Because of the differences in their skill representations, which determine how the given example is being modeled, LfD algorithms exhibit distinct reproduction behaviors. Some common behavior for reproductions include quickly converging and following the given demonstration, or preserving the shape of the demonstration from the given initial point.

\begin{figure*}[h]
\centering
\includegraphics[trim=0 0em 0 0, clip, width=0.9\textwidth]{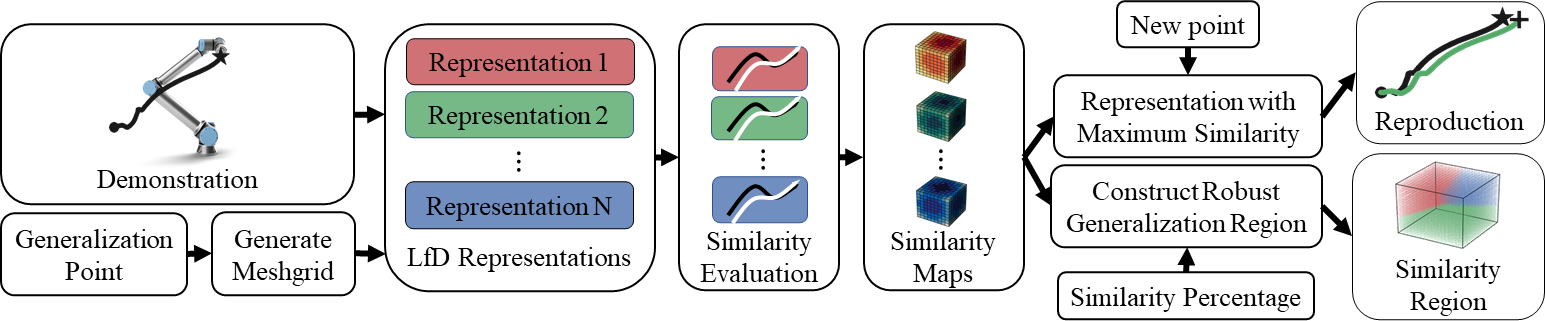}
\caption{\small{
A flow diagram illustrating the proposed similarity-aware multi-representational LfD (SAMLfD) framework.}} \label{flowchart}
\end{figure*}

We show that when generalizing a learned skill over different boundary conditions (e.g., initial point), a single LfD representation cannot preserve a consistent similarity, and instead we propose a similarity-aware LfD framework consisting of multiple representations together with a similarity metric (see Fig. \ref{flowchart}). The contributions of this framework are an
\begin{enumerate*}[label=(\roman*)]
    \item better generalization capabilities through understanding representation performance across a generalization space as well as
    \item providing information to the user about the generalization ability of LfD representations.
\end{enumerate*}
Given a single demonstration of a skill, our framework evaluates the outcomes of different representations using a specified similarity metric and provides the user with (a) the most similar reproduction of the skill for a given new condition including initial or final points, and (b) a similarity region, which contains information on each representation's ability to generalize over different points in space. For an unforeseen situation (e.g., new initial point), the proposed framework generalizes a learned skill with the highest similarity by considering the performance of each individual representation.

We validate our framework in three simulated and four real-world experiments using a physical 6-DOF robot. Our framework is not limited to a specific set of LfD representations or a particular similarity metric. However, we show that most metrics favor a specific reproduction behavior (e.g., converging to the demonstration vs. preserving the shape of the trajectory) which means they are implicitly biased towards a specific representation. To categorize similarity metrics according to their biases towards different reproduction behaviors, we have evaluated 26 skills with 11 metrics for a total of 286 simulated experiments.

\begin{figure*}[ht]
\centering
\includegraphics[trim=0 0em 0 0, clip, width=0.98\textwidth]{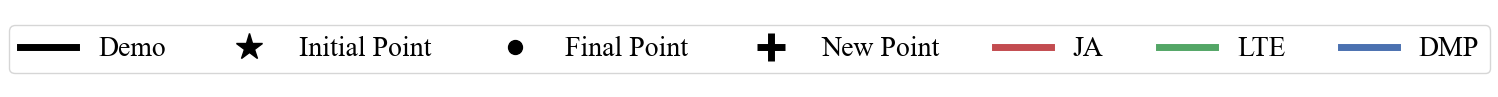}
\vspace{-1em}
\caption{\small{
All figures and tables presented in this paper follow this legend.
}}\label{legend}
\end{figure*}

\section{Related Work}
\label{sec:rw}
Extensive research has been done on LfD and its representations \cite{ravichandar2020recent}. There exists a wide variety of representations, each with a unique method for interpreting and encoding demonstrations and crafting a reproduction. For instance, some existing LfD representations encode demonstrations using statistical~\cite{hertel2021TLFSD} or probabilistic~\cite{ProMP} methods, while others use  geometric models~\cite{TLGC} or dynamical systems~\cite{pastorDMP2009, LASA}. Some representations attempt to combine these methods in order to improve performance~\cite{huang2019kernelized}. A representation's encoding method implicitly defines its performance and significantly affects its reproduction and generalization capabilities. As a result different representations have shown to perform differently on various skills~\cite{GT_dataset}. Number of required demonstrations is another factor that contributes to the behavior and performance of the LfD representations. Some LfD representations~\cite{LTE,JA,pastorDMP2009} rely on a single example of the skill. Although this might be convenient for the end-user, one demonstration provides minimal information about variance. Other LfD representations benefit from the spatial and temporal variance provided by a set of multiple demonstrations~\cite{TLGC, ProMP}. However, end-users with different levels of experience might not be able to provide multiple demonstrations of a skill that are diverse enough, cover the demonstration space, and satisfy all the constraints. Consequently, the quality of demonstrations from users with varying experience affects the performance of the LfD representations~\cite{GT_dataset}. 

Several approaches have been developed to address the discussed issues. Confidence-Based Learning from Demonstration~\cite{confidence2007} attempts to improve robot's generalization capability by measuring the uncertainty of reproducing from a novel initial point and requesting a new demonstration from the user in uncertain areas. Unlike our framework, this method is incremental and requires additional demonstrations from the user. Similarly, preference-based approaches improve robot skill generalization by querying the user iteratively for each reproduction which could be tedious and user-dependent~\cite{palan2019learning}.

In some approaches, the LfD representation has been equipped with a skill refinement strategy that allows the user to refine the reproduced trajectory through kinesthetic corrections~\cite{TLGC} or by assigning an importance value to each demonstration~\cite{argall2009automatic}. Providing improving kinesthetic corrections or effective weight values could be complicated for inexperienced users. Multi-Coordinate Cost Balancing (MCCB)~\cite{MCCB} transforms the given set of demonstrations into multiple differential coordinates and optimizes across coordinates to generate an improved reproduction of the skill. While MCCB focuses on finding the most important features that should be preserved, our framework can use similarity metrics that favor other features as well. Alternatively, Deep Reinforcement Learning has been used to improve skill generalization in novel situations~\cite{generalization_DMF, andrychowicz2017hindsight}. Although these approaches enable the robot to find optimal reproductions by exploring the state-space, a large amount of data is required for training the model and adapting to new situations.

\begin{figure}[ht]
\centering
\includegraphics[trim=0 0em 0 0, clip, width=0.35\textwidth]{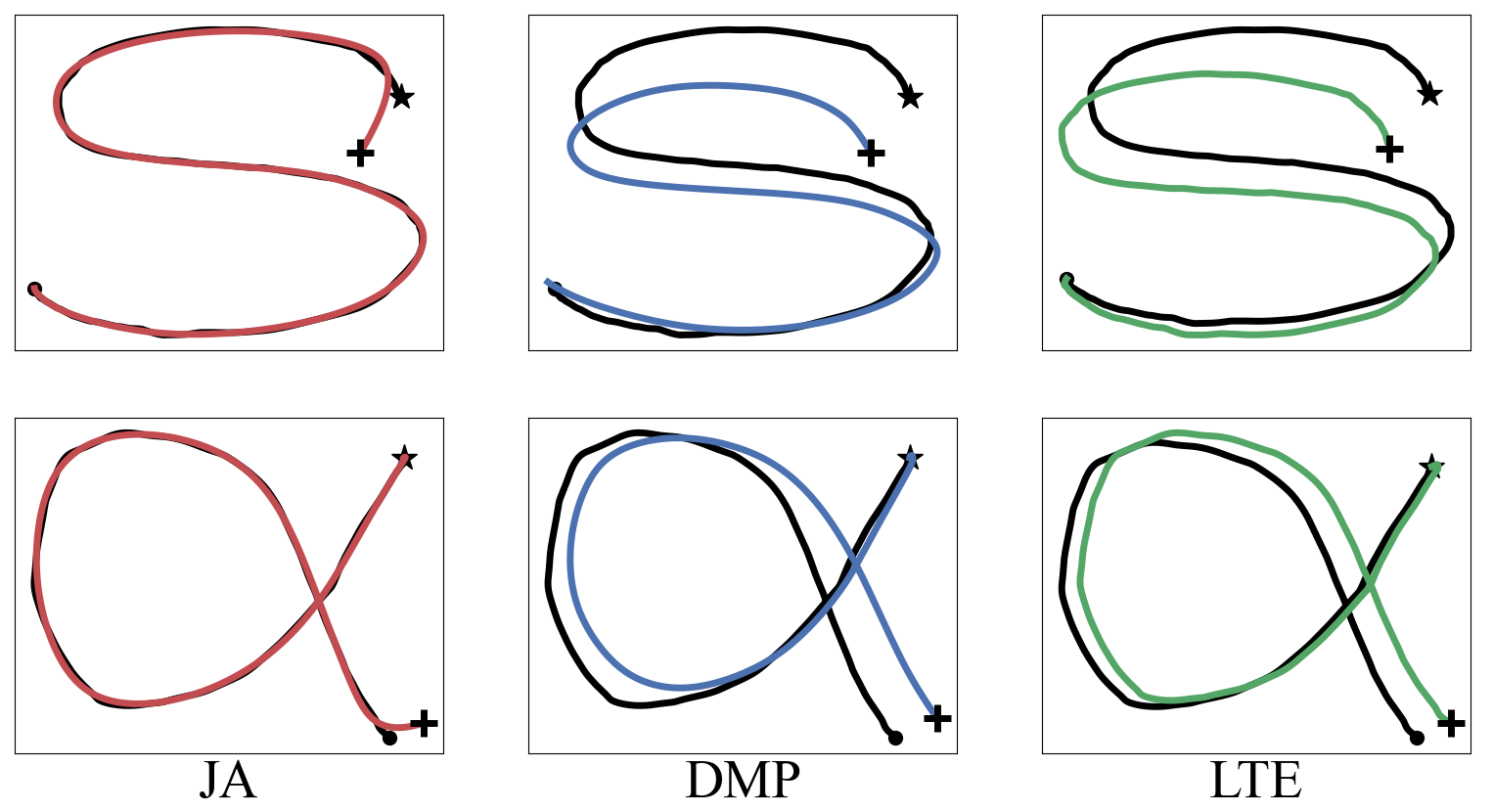}
\caption{\small{
Reproductions of letters $S$ and $\alpha$ using three LfD representations with different reproduction behaviors ranging from immediate convergence in JA to preserving shape in LTE.
}}\label{behavior}
\end{figure}

\section{Background}
\label{sec:bg}

In this section, we provide a brief description of the LfD representations included in our multi-representational LfD framework, namely, Laplacian Trajectory Editing~\cite{LTE}, Jerk Accuracy Model~\cite{JA}, and Dynamic Movement Primitives~\cite{pastorDMP2009}. It should be noted that our framework is not restricted to these representations and any existing (single-demonstration) LfD representation can be included, these representations were selected due to their different reproduction behavior. The behavior of these representations can be seen in two experiments in Fig.~\ref{behavior}. Note that all representations rely on a single demonstration being provided as $\bm{X} = [\bm{x}(0),\bm{x}(1),...,\bm{x}(T)]^{\top} \in \mathbb{R} ^{T \times n}$ where $\bm{x}(t) \in \mathbb{R}^n$ denotes a discrete $n$-dimensional sample at time index $t$.

\subsection{Laplacian Trajectory Editing}
Laplacian Trajectory Editing (LTE) is an LfD representation that uses the Laplacian differential coordinate to reproduce and generalize a demonstrated skill~\cite{LTE}. Given a demonstration LTE represents the path as a graph and transforms it to the Laplacian coordinate using the discrete Laplace-Beltrami operator $\bm{L}$ as $\boldsymbol{\Delta} = \mathbf{LX}$. Additional constraints of the trajectory (e.g., a fixed goal point) are specified in the form $\mathbf{\bar{X}X} = \mathbf{\bar{C}}$ resulting in the following equation system
\begin{equation}
    \left[\begin{matrix}\mathbf{L} \\ \mathbf{\bar{X}} \end{matrix}\right]\mathbf{X}_r = \left[\begin{matrix}\boldsymbol{\Delta} \\ \mathbf{\bar{C}} \end{matrix}\right], %\nonumber
\end{equation}
which can be solved to find the reproduction $\mathbf{X}_r \in \mathbb{R} ^{T \times n}$ using matrix inversion as
\begin{equation}
    \mathbf{X}_r = \left[\begin{matrix}\mathbf{L} \\ \mathbf{\bar{X}} \end{matrix}\right]^+ \left[\begin{matrix}\boldsymbol{\Delta} \\ \mathbf{\bar{C}} \end{matrix}\right]. %\nonumber
\end{equation}
LTE preserves the curvature of the demonstrated trajectory while satisfying the given constraints. As a result, the reproduced trajectory $\mathbf{X}_r$ does not necessarily converge to and follow the demonstration $\textbf{X}$. 

\subsection{Jerk Accuracy Model}
Jerk Accuracy (JA) model is an optimization-based LfD representation that aims at maximizing smoothness and geometrical invariance of a template model~\cite{JA}. Given a demonstration the JA representation generates a reproduction through optimization of a set of objectives $I$ subject to a set of constraints $\vect{C}$ as:
\begin{equation}
    \mathrm{min} \; \mathit I(\vect{X}) \;\;\;\; \mathrm{\tab subject \; to} \; \vect{C}, \label{eq:ja} %\nonumber
\end{equation}
where the objectives form a trade-off between smoothness (the minimization of jerk) and accuracy using the following functional:
\begin{equation}
     I_{\vect{X}, T, \lambda} = \int_0^T  \abs{\vect{\dddot{X}}_r}^2 \, dt + \lambda^6 \int_0^T  \abs{\vect{X}_r - \vect{X}}^2 \, dt, %\nonumber
 \end{equation}
which uses the Lagrange multiplier $\lambda$ to modulate the weighting between smoothness and accuracy. The optimization problem in~\eqref{eq:ja} is then solved as a boundary value problem. JA is especially suited to finding smooth reproductions of noisy trajectories while satisfying the constraints. JA reproductions from a new generalization point quickly converge to the demonstration.

\subsection{Dynamic Movement Primitives}
Dynamic Movement Primitives (DMPs)\cite{pastorDMP2009} reproduce a given demonstration using a linear dynamical system perturbed by a nonlinear forcing term as follows:
\begin{align}
    \tau \dot{\vect{v}} &= K(\vect{g} - \vect{x}) - D\vect{v} - K(\vect{g} - \vect{x}_0)s + K\vect{f}(s) \label{equ:dmp1} \\  
    \tau \dot{\vect{x}} &= \vect{v}, %\nonumber
    \label{equ:dmp2}
\end{align}
\noindent where $\vect{x}$ and $\vect{x}$ are the position and velocity vectors, $\vect{g}$ is the end-point, $\vect{x}_0$ is the initial point, $\tau$ is a temporal scaling factor, $K$ and $D$ are constants chosen such that the system is critically damped, and $\vect{f}$ is a non-linear function which is learned using linear weighted regression to allow the generation of arbitrarily complex movements. The transformation system defined in \eqref{equ:dmp1} and \eqref{equ:dmp2} is time-independent and driven by a canonical system formulated as $\tau \dot{s} = - \alpha_s s$ where $s(t)$ is the phase variable which monotonically changes from 1 to 0 over time and $\alpha_s$ is a predefined constant. DMPs can reproduce trajectories that converge to the goal $\vect{g}$ while being spatial and temporal invariant and are robust against perturbations. 

\section{Methodology}
\label{sec:method}

In this section, we propose a novel framework called Similarity-Aware Multi-representational Learning from Demonstration (SAMLfD) that improves robot skill reproduction and generalization capabilities by relying on finding the best LfD representation to satisfy a given similarity metric (with high-level flow diagram shown in Fig.~\ref{flowchart}). 

It should be noted that while expert users may already have inherent knowledge of which LfD representation is needed to properly generalize a task~\cite{GT_dataset}, our framework is intended to enable end-users to teach robots new skills regardless of their level of experience. In this paper, we focus on the use of single-demonstration LfD representations, which only are simpler to use, and leave the use of multi-demonstration representations to future work. Additionally, we use common similarity metrics and focus on primitive tasks, although SAMLfD is able to provide the user with a high-level overview of generalization across a region for any task.

Given a demonstration and a point of interest (can be initial or final point), SAMLfD first generates a meshgrid--a set of points distributed uniformly around the point of interest with size and point density chosen by the user. In a coarse grid, the framework evaluates skill generalization over a larger task space with less points  thus losing information continuity. In a dense grid, however, the amount of information obtained increases at the cost of computational complexity. After constructing the meshgrid, SAMLfD generates one reproduction of the skill using each LfD representation from each point in the grid. Next, SAMLfD evaluates all reproductions using the given similarity metric. For each representation, the normalized similarity values form a similarity map (the heatmap cubes in Fig.~\ref{flowchart}). These maps are combined by selecting the greatest similarity reproduction at each point on the meshgrid and feeding the data to a classifier to understand how a representation performs across the generalization space. The trained classifier is used to classify generalization points that were not included in the initial grid. In this paper, we have used a Support Vector Machine (SVM) classifier variant known as C-Support Vector Classification which supports multi-class classification, but the proposed framework does not depend on this specific classifier. The combined map displays similarity regions where a specific representation outperforms other ones.

The next steps of SAMLfD are based on the user's decision. Should the user want a reproduction at a specific point, the framework returns a reproduction produced by the representation with the highest similarity at that point (the reproduction shown in Fig.~\ref{flowchart}). Alternatively, SAMLfD can show the user the generalization regions of the given representations (the similarity cube shown in Fig.~\ref{flowchart}). By default, the classifier uses all data and generates similarity regions based on the best reproductions. However, should the user only want to see regions within which the reproductions can perform the skill relatively well (e.g., with 75\% similarity), SAMLfD can construct robust generalization regions which obtain this performance and provide this information to the user. The user can utilize this information to make informed decisions about the task and region in the generalization space for which the skill can be adequately generalized.

The proposed framework is intended to be highly adjustable and can be adapted to any situation based on the user's inputs. Given demonstrations can be in joint space or task space. The framework can be equipped with various LfD representations that rely on a single demonstration as well as different similarity metrics. Since most similarity metrics are not bounded on one end, the framework assigns the reproduction with the greatest raw similarity a value of 1 and maps all other values to the range $[0,1]$. Unbounded similarity metrics means that factors such as the extent of the generalization space can affect the normalized similarity values of reproduction. Additionally, the generalization ability of the framework is limited to the generalization ability of the representations used.\footnote{Implementation: \url{https://github.com/brenhertel/SAMLfD}}

\begin{figure}[ht]
\centering
\includegraphics[trim=0 0em 0 0, clip, width=\columnwidth]{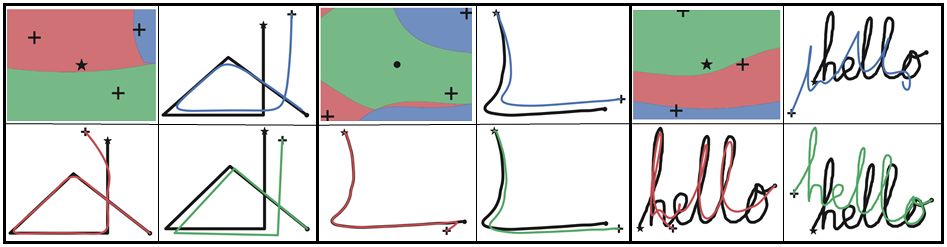}
\caption{\small{
Generalization of three 2D skills using SAMLfD with three representations (See Section~\ref{subsec:simulated_exps} for details).
}}\label{2d_demos}
\end{figure}

\section{Experiments}
\label{sec:exp}

We have evaluated the performance of our framework in 3 simulated and 4 sets of real-world experiments. For the simulated experiments, a user has provided each demonstration by interacting with a GUI through a pointing device (e.g., mouse). For the real-world experiments, demonstrations are provided via kinesthetic teaching on a 6-DOF UR5e manipulator. We have recorded trajectories in task-space and applied a noise filtering and resampling process to the raw data. In our multi-representational framework, we have used JA~\cite{JA}, LTE~\cite{LTE}, and DMP~\cite{pastorDMP2009} LfD representations, with the Fr\'echet distance~\cite{frechet_computing} as the similarity metric. It should be noted that LTE requires no parameters and for consistency we kept the parameters of JA and DMP fixed for all experiments. In 3D experiments, we used spherical linear interpolation of quaternions~\cite{slerp} for generating orientation trajectories before execution. Reproductions performed on the manipulator are computed a priori and executed using a low-level controller. See Fig.~\ref{legend} for color and marker conventions.

\begin{figure*}[ht]
\centering
\includegraphics[trim=0 0em 0 0, clip, width=0.8\textwidth]{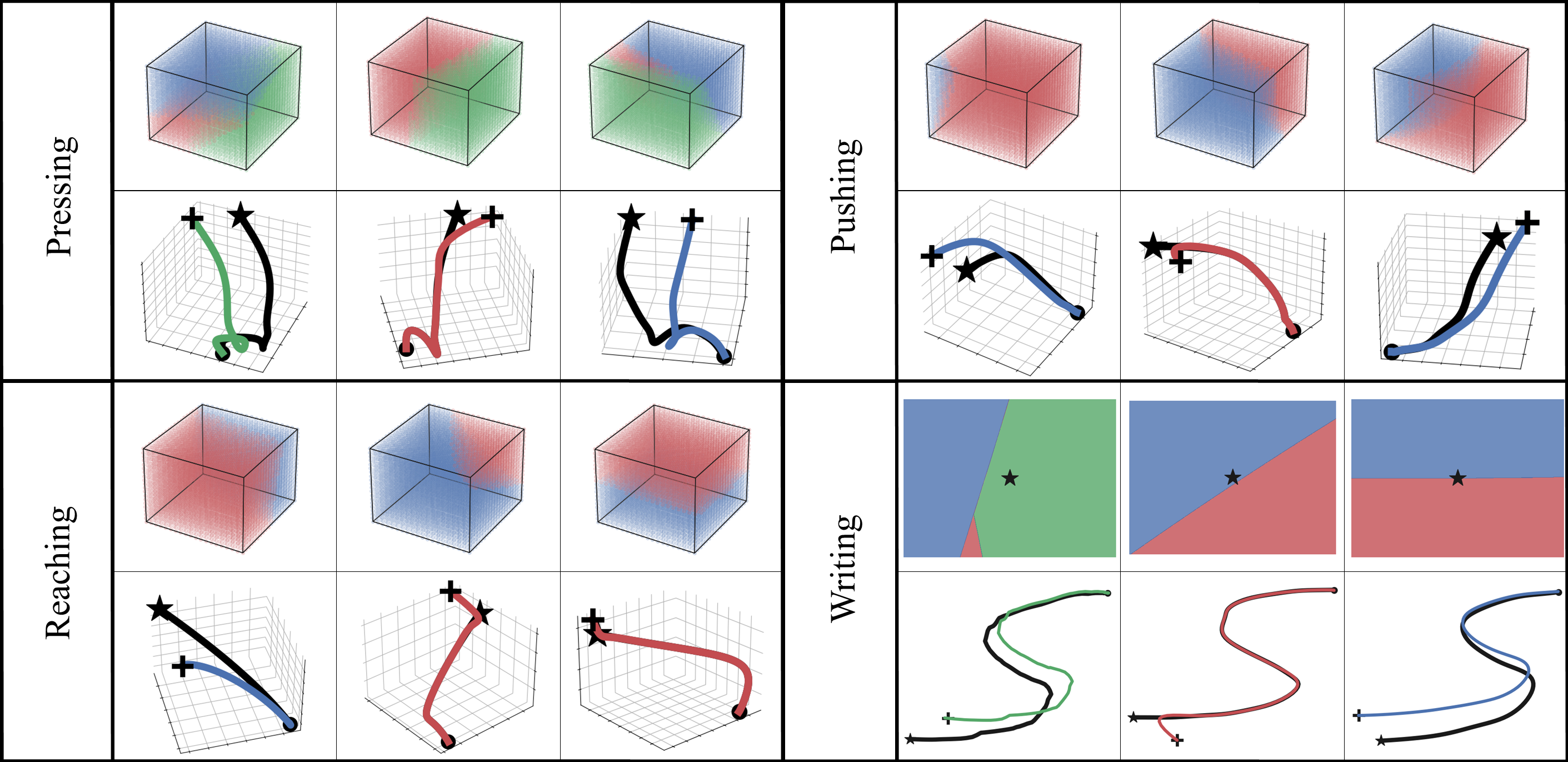}
\caption{Generalization of four real-world skills using SAMLfD. For each skill, the experiment was repeated three times with a different input demonstration. The similarity regions can be seen above each test (see Section~\ref{subsec:real_world_exps} for details).
}\label{GT_data}
\end{figure*}

\subsection{Simulated Experiments}
\label{subsec:simulated_exps}

Fig.~\ref{2d_demos} shows results for three simulated experiments using 2D demonstrations of a ``Straight Alpha,'' the ``LShape,'' and the word ``hello''. For each experiment, the obtained similarity region for a given generalization point is depicted on the top left corner as well as reproductions of the skill for new conditions selected randomly from different regions. In all the experiments at least one region per representation has appeared. In some cases, we expect to see less than three regions become present, which can happen because either the similarity metric is biased toward a specific representation or two representations produce very similar outcomes.  

The set of subplots on the left of Fig.~\ref{2d_demos} show results for the generalization of the ``Straight Alpha'' shape over the randomly selected initial points. Although these reproductions are the most similar reproductions of the skill from the selected points, it can be seen that the LTE reproduction preserves the sharp corners of the demonstration, while the JA and DMP produce rounded corners. The next experiment, shown in the middle of Fig.~\ref{2d_demos}, generalizes the ``LShape'' demonstration (from the LASA dataset~\cite{LASA}) over its endpoint. It can be seen that SAMLfD suggests the use of LTE representation for the most of the generalization region. Here, LTE reproductions look most similar to the original demonstration. The set of subplots on the right of Fig.~\ref{2d_demos} show results for the generalization of the word ``hello'' over its initial point. This trajectory is more complex than previous examples. A large part of the generalization region is covered by the LTE representation as reproductions maintain the shape of the trajectory despite the complexity, and ``hello'' can clearly be seen in the reproduction. The rest of the generalization space is mostly covered by JA reproductions which are somewhat readable but struggle to preserve high-jerk features like the tight loops in letters such as ``o'' or ``e.''  The DMP reproduction also struggles to reproduce a readable trajectory. It should be noted that due to the complexity of the trajectory, there is no single tuning of DMP which allows for visually appealing reproductions across the generalization region. However, we used parameters that create a section in which DMP outperforms JA and LTE according to the Fr\`echet distance metric. 

\subsection{Real-World Experiments}
\label{subsec:real_world_exps}

We also evaluated the proposed framework in real-world experiments using the dataset provided in~\cite{GT_dataset} that contains demonstrations of four 3D skills: pressing, pushing, reaching, and writing. For each skill, the experiment was repeated three times, each time with a different input demonstration. In these experiments, we consider only generalization over the initial point. Fig.~\ref{GT_data} shows the similarity region constructed by SAMLfD for the given demonstration in the top row and the demonstration and the corresponding best reproduction in the second row for each skill. It should be noted that similarity regions and reproductions have not necessarily been plotted from the same point of view. Reproductions were selected through a GUI (see the accompanying video\footnote{Video available at: \url{https://youtu.be/LZW3MWrNFSI}}). For the pressing and writing skills, all three representations appear in the similarity map. These skills can be accomplished either by converging to the demonstration or preserving the shape of the given demonstration. For the pushing and reaching skills, JA and DMP are the representations found to reproduce with greatest similarity, that makes the presence of the LTE representation minimal. For these skills, convergence to the trajectory is more important than preserving the shape of the trajectory, which would lead JA and DMP to be better representations for these skills. The accompanying video shows the generalization of a reaching skill from an arbitrary demonstration, the user interface for selecting a new initial point, and execution of the skill from three points selected from different regions using a UR5e manipulator arm. 

\begin{figure}[t]
\centering
\includegraphics[trim=0 0em 0 0, clip, width=0.70\columnwidth]{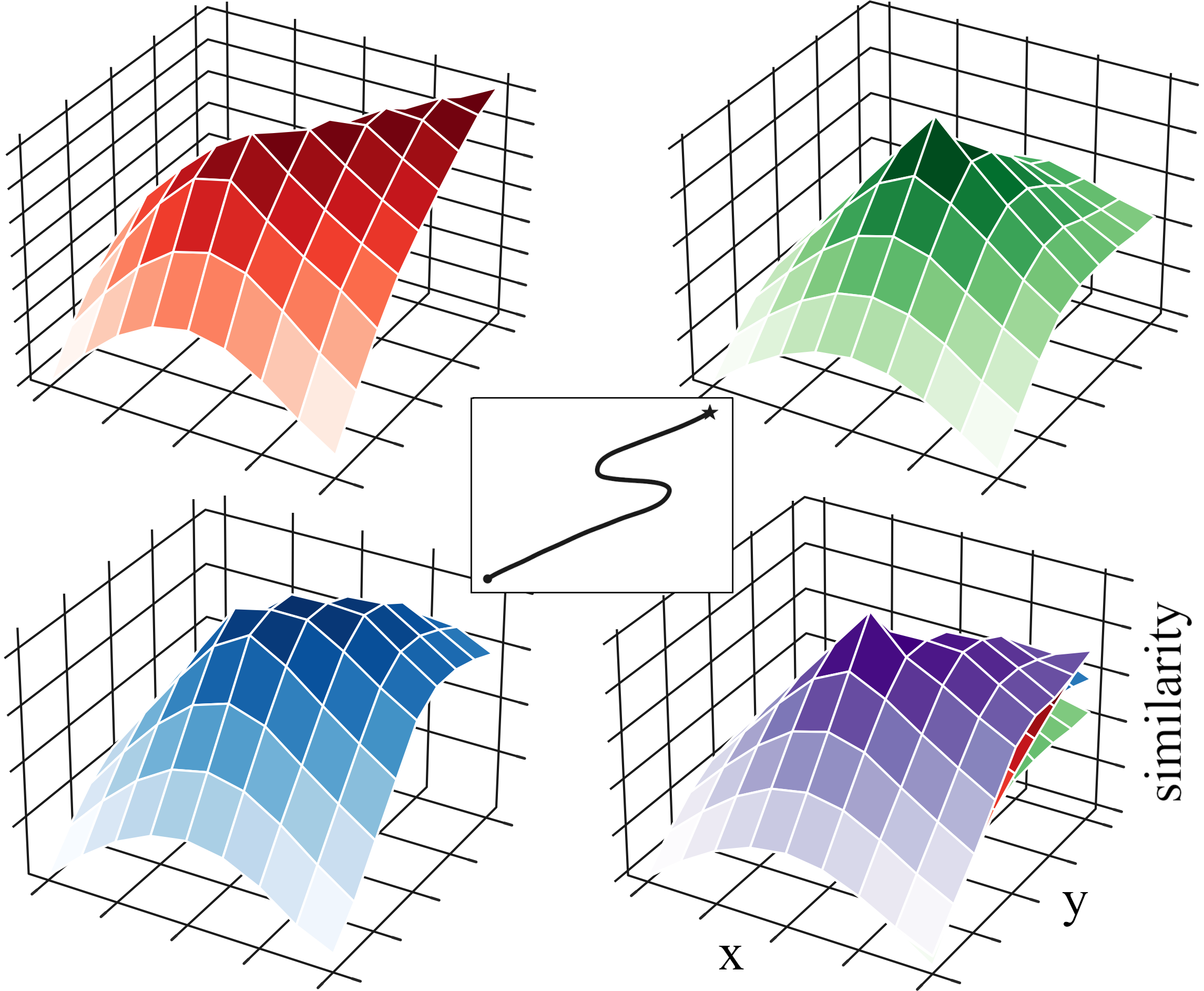}
\caption{\small{
Surface plots of similarity values for a writing skill (center). Similarity maps of each representation are plotted individually and combined to construct the highest similarity map (shown on bottom right) which outperforms surfaces of other representations.
}}\label{surface_comparison}
\end{figure}

\begin{figure*}[ht]
\centering
\includegraphics[trim=0 0em 0 0, clip, width=0.90\textwidth]{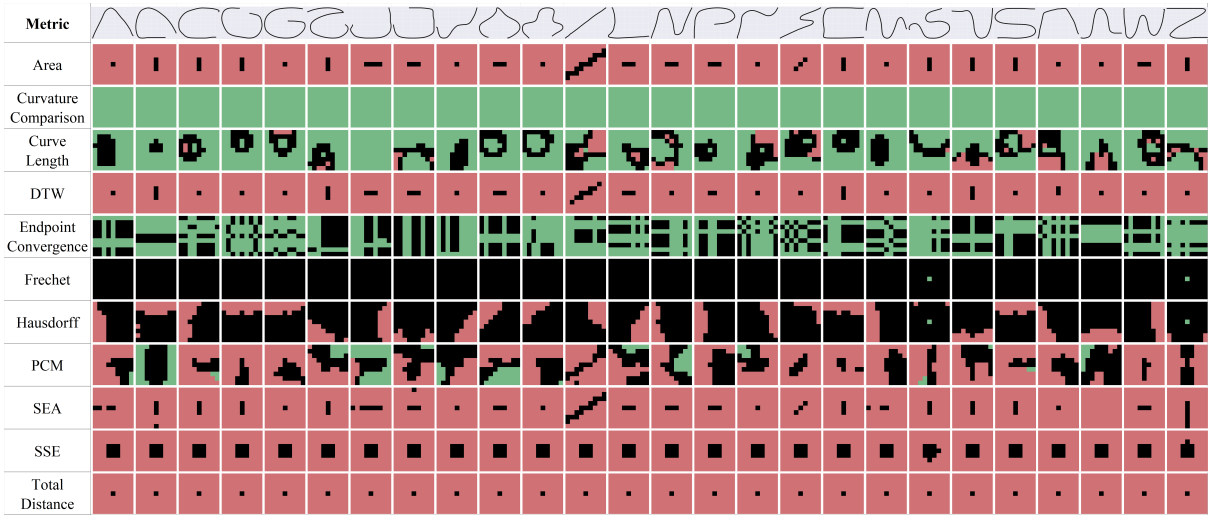}
\caption{\small{
A visual representation of how metrics responded to different representations using the LASA dataset. Red is where metric favors convergence, green is where metric favors shape preservation, and black is where metric can favor either behavior (inconclusive).}}\label{metric_maps}
\end{figure*}

\subsection{Comparison to Individual LfD Representations}
\label{subsec:representation_comparison}

We also show that the combination of representations to create a highest similarity reproduction leads to more similar reproductions across the generalization region. Fig.~\ref{surface_comparison} shows similarity maps obtained from the three LfD representations for a 2D shape from the LASA dataset~\cite{LASA}. Each surface depicts similarity over a meshgrid around the initial point of the given demonstration. In this experiment, we used Haussdorff distance~\cite{hausdorff} as the similarity metric. While the three representations are all able to generalize this writing skill, the SAMLfD surface represents highest similarity at each point to generate a reproduction. To measure the difference in similarity across the generalization space, we calculated accumulated similarity differences at each point $ij$ on the meshgrid as $\delta_{\text{rep}} = \sum_{ij} S_{\text{SAMLfD}}^{ij} - S_{\text{rep}}^{ij}$ for $\text{rep} \in \{\text{JA}, \text{LTE}, \text{DMP}\}$. The similarity differences in this experiment were found to be $\delta_{\text{JA}}=5.19$, $\delta_{\text{LTE}}=2.93$, and $\delta_{\text{DMP}}=0.99$. This shows that our multi-representational framework is able to achieve a higher similarity result across a generalization space than any individual representation.

\section{Quantitative Metric Evaluations}
\label{subsec:metric_eval}

In general, LfD representations have been designed to mimic the given demonstration by satisfying the desired constraints while preserving important properties of the trajectory (e.g., shape). Different representations, however, place different importance on these objectives. For instance, JA prefers immediate convergence to the demonstrated trajectory, while LTE focuses on preserving its curvature (see Fig.~\ref{behavior}). As a result, similarity metrics could be biased towards either of these behaviors. 

In this section, we study metric bias and perform experiments to categorize a set of known metrics into two groups: (a) metrics that favor \emph{convergence} and (b) metrics that favor \emph{shape preservation}. The first group measures similarity between two curves according to how much they converge, while the second group measures shape feature similarities (e.g., curvature). For this study, we configured our multi-representational LfD framework using the LTE and JA representations. While the former favors preserving curvature, the latter prefers immediate convergence with minimum jerk. We used this setup to evaluate 11 different metrics (as listed in Table~\ref{metric_nums}) on all 26 skills from the LASA dataset~\cite{LASA} for a total of 286 experiments. For each evaluation, a meshgrid of 81 points centered on the initial point of the demonstration was used.

Fig.~\ref{metric_maps} illustrates the obtained similarity maps for each metric. The color of each point (red or green) indicates the representation that resulted in a reproduction with greatest similarity at that point. If the similarity value of a point for both reproductions is within 10\% of the other, it was deemed inconclusive and shown in black. Combined across 26 shapes, this gives 2,106 data points for each metric. One color dominating the maps for a metric means that metric shows a bias toward that representation and favors its behavior. Table~\ref{metric_nums} categorizes each metric according to the observed metric bias towards different representations. Each value is calculated by the total number of times a representation was selected over total number of datapoints. For instance, DTW measured 97.58\% JA appearances and 2.42\% inconclusive appearances, meaning that DTW favors the converging behavior. Fig.~\ref{metric_maps} shows that some metrics are extremely biased towards one type of behavior. For instance, the curvature comparison metric favors  shape preservation, while the total distance favors  trajectory convergence. In the experiments in the previous section, we used Fr\'echet distance, which according to the results from this section, does not exhibit a bias towards either behavior. This also means that Fr\'echet distance may not be preferred for evaluating a specific skill. For example, with the reaching skill, a user may prefer total distance rather than Fr\'echet distance, because it is biased towards trajectory convergence. One of the advantages of our framework is that it allows the user to select their metric of choice to evaluate and generalize the skill accordingly. While less experienced users can rely on a pre-selected similarity metric such as the Fr\'echet distance, more experienced users can benefit from selecting a specific metric according to their task objectives.

\begin{table}[h]
\centering
\footnotesize
\caption{\small{
Numerical evaluation of the data shown in Fig.~\ref{metric_maps}. Each metric has been categorized as whether it prefers convergence (JA), shape preservation (LTE), or is inconclusive (either).}}\label{metric_nums}
\begin{tabular}{c|c|c|c|c}
\toprule
\textbf{Metric} & \textbf{JA} & \textbf{LTE} & \textbf{Inconclusive} & \textbf{Decision} \\ \midrule
Area                                                          & {\color[HTML]{CF7175} 96.44\%}                                                       & {\color[HTML]{77B986} 0.00\%}                                                         & 3.56\%                                     & {\color[HTML]{CF7175} JA}  \\ \hline
\begin{tabular}[c]{@{}c@{}}Curvature \\ Comparison\end{tabular} & {\color[HTML]{CF7175} 0.00\%}                                                        & {\color[HTML]{77B986} 100.00\%}                                                       & 0.00\%                                     & {\color[HTML]{77B986} LTE} \\ \hline
\begin{tabular}[c]{@{}c@{}}Curve Length\end{tabular}           & {\color[HTML]{CF7175} 7.36\%}                                                        & {\color[HTML]{77B986} 65.34\%}                                                        & 27.30\%                                    & {\color[HTML]{77B986} LTE} \\ \hline
DTW                                                          & {\color[HTML]{CF7175} 97.58\%}                                                       & {\color[HTML]{77B986} 0.00\%}                                                         & 2.42\%                                     & {\color[HTML]{CF7175} JA}  \\ \hline
\begin{tabular}[c]{@{}c@{}}Endpoint \\ Convergence\end{tabular}   & {\color[HTML]{CF7175} 0.00\%}                                                        & {\color[HTML]{77B986} 58.50\%}                                                        & 41.50\%                                    & {\color[HTML]{77B986} LTE} \\ \hline
Fr\'echet                                                    & {\color[HTML]{CF7175} 0.00\%}                                                        & {\color[HTML]{77B986} 0.09\%}                                                         & 99.91\%                                    & Either                     \\ \hline
Hausdorff                                                      & {\color[HTML]{CF7175} 24.60\%}                                                       & {\color[HTML]{77B986} 0.09\%}                                                         & 75.31\%                                    & Either                     \\ \hline
PCM                                                    & {\color[HTML]{CF7175} 59.07\%}                                                       & {\color[HTML]{77B986} 9.26\%}                                                         & 31.67\%                                    & {\color[HTML]{CF7175} JA}  \\ \hline
SEA                                                              & {\color[HTML]{CF7175} 95.92\%}                                                       & {\color[HTML]{77B986} 0.00\%}                                                         & 4.08\%                                     & {\color[HTML]{CF7175} JA}  \\ \hline
SSE                                                               & {\color[HTML]{CF7175} 88.75\%}                                                       & {\color[HTML]{77B986} 0.00\%}                                                         & 11.25\%                                    & {\color[HTML]{CF7175} JA}  \\ \hline
\begin{tabular}[c]{@{}c@{}}Total \\ Distance\end{tabular}         & {\color[HTML]{CF7175} 98.77\%}                                                       & {\color[HTML]{77B986} 0.00\%}                                                         & 1.23\%                                     & {\color[HTML]{CF7175} JA} \\
\bottomrule
\end{tabular}
\end{table}

%%%%%%%%%%%%%%%%%%%%%%%%%%%%%%%%%%%%%%%%%%%%%%%%%%%%%%%%%%%%
% \section{Acknowledgements}
%%%%%%%%%%%%%%%%%%%%%%%%%%%%%%%%%%%%%%%%%%%%%%%%%%%%%%%%%%%%

\addtolength{\textheight}{-12cm} 

\bibliographystyle{IEEEtran}
\bibliography{references}

\begin{thebibliography}{10}
\providecommand{\url}[1]{#1}
\csname url@rmstyle\endcsname
\providecommand{\newblock}{\relax}
\providecommand{\bibinfo}[2]{#2}
\providecommand\BIBentrySTDinterwordspacing{\spaceskip=0pt\relax}
\providecommand\BIBentryALTinterwordstretchfactor{4}
\providecommand\BIBentryALTinterwordspacing{\spaceskip=\fontdimen2\font plus
\BIBentryALTinterwordstretchfactor\fontdimen3\font minus
  \fontdimen4\font\relax}
\providecommand\BIBforeignlanguage[2]{{%
\expandafter\ifx\csname l@#1\endcsname\relax
\typeout{** WARNING: IEEEtran.bst: No hyphenation pattern has been}%
\typeout{** loaded for the language `#1'. Using the pattern for}%
\typeout{** the default language instead.}%
\else
\language=\csname l@#1\endcsname
\fi
#2}}

\bibitem{ravichandar2020recent}
H.~Ravichandar, A.~S. Polydoros, S.~Chernova, and A.~Billard, ``Recent advances
  in robot learning from demonstration,'' \emph{Annual Review of Control,
  Robotics, and Autonomous Systems}, vol.~3, 2020.

\bibitem{hertel2021TLFSD}
B.~Hertel and S.~R. Ahmadzadeh, ``Learning from successful and failed
  demonstrations via optimization,'' in \emph{IEEE/RSJ International Conference
  on Intelligent Robots and Systems (IROS)}.\hskip 1em plus 0.5em minus
  0.4em\relax IEEE, 2021.

\bibitem{ProMP}
A.~Paraschos, C.~Daniel, J.~R. Peters, and G.~Neumann, ``Probabilistic movement
  primitives,'' in \emph{Advances in neural information processing systems},
  2013, pp. 2616--2624.

\bibitem{TLGC}
S.~R. Ahmadzadeh and S.~Chernova, ``Trajectory-based skill learning using
  generalized cylinders,'' \emph{Frontiers in Robotics and AI}, vol.~5, p. 132,
  2018.

\bibitem{pastorDMP2009}
P.~Pastor, H.~Hoffmann, T.~Asfour, and S.~Schaal, ``Learning and generalization
  of motor skills by learning from demonstration,'' in \emph{2009 IEEE
  International Conference on Robotics and Automation}.\hskip 1em plus 0.5em
  minus 0.4em\relax IEEE, 2009, pp. 763--768.

\bibitem{LASA}
S.~M. Khansari-Zadeh and A.~Billard, ``Learning stable nonlinear dynamical
  systems with gaussian mixture models,'' \emph{IEEE Transactions on Robotics},
  vol.~27, no.~5, pp. 943--957, 2011.

\bibitem{huang2019kernelized}
Y.~Huang, L.~Rozo, J.~Silv{\'e}rio, and D.~G. Caldwell, ``Kernelized movement
  primitives,'' \emph{The International Journal of Robotics Research}, vol.~38,
  no.~7, pp. 833--852, 2019.

\bibitem{GT_dataset}
R.~M. Asif, C.~Dephne, W.~Jacob, C.~Vivian, S.~R. Ahmadzadeh, and C.~Sonia,
  ``Benchmark for skill learning from demonstration: Impact of user experience,
  task complexity, and start configuration on performance,'' in \emph{Robotics
  and Automation ({ICRA}), {IEEE} International Conference on}.\hskip 1em plus
  0.5em minus 0.4em\relax Paris, France: {IEEE}, May 2020, pp. 7561--7567.

\bibitem{LTE}
T.~Nierhoff, S.~Hirche, and Y.~Nakamura, ``Spatial adaption of robot
  trajectories based on laplacian trajectory editing,'' \emph{Autonomous
  Robots}, vol.~40, no.~1, pp. 159--173, 2016.

\bibitem{JA}
Y.~Meirovitch, D.~Bennequin, and T.~Flash, ``Geometrical invariance and
  smoothness maximization for task-space movement generation,'' \emph{IEEE
  Transactions on Robotics}, vol.~32, no.~4, pp. 837--853, 2016.

\bibitem{confidence2007}
S.~Chernova and M.~Veloso, ``Confidence-based policy learning from
  demonstration using gaussian mixture models,'' in \emph{Proceedings of the
  6th international joint conference on Autonomous agents and multiagent
  systems}, 2007, pp. 1--8.

\bibitem{palan2019learning}
M.~Palan, N.~C. Landolfi, G.~Shevchuk, and D.~Sadigh, ``Learning reward
  functions by integrating human demonstrations and preferences,'' in
  \emph{Proceedings of Robotics: Science and Systems (RSS)}, June 2019.

\bibitem{argall2009automatic}
B.~D. Argall, B.~Browning, and M.~Veloso, ``Automatic weight learning for
  multiple data sources when learning from demonstration,'' in \emph{2009 IEEE
  International Conference on Robotics and Automation}.\hskip 1em plus 0.5em
  minus 0.4em\relax IEEE, 2009, pp. 226--231.

\bibitem{MCCB}
H.~Ravichandar, S.~R. Ahmadzadeh, M.~A. Rana, and S.~Chernova, ``Skill
  acquisition via automated multi-coordinate cost balancing,'' in \emph{2019
  International Conference on Robotics and Automation (ICRA)}.\hskip 1em plus
  0.5em minus 0.4em\relax IEEE, 2019, pp. 7776--7782.

\bibitem{generalization_DMF}
T.~Wang, H.~Zhang, W.~Q. Toh, H.~Zhu, C.~Tan, Y.~Wu, Y.~Liu, and W.~Jing,
  ``Efficient robotic task generalization using deep model fusion reinforcement
  learning,'' in \emph{2019 IEEE International Conference on Robotics and
  Biomimetics (ROBIO)}.\hskip 1em plus 0.5em minus 0.4em\relax IEEE, 2019, pp.
  148--153.

\bibitem{andrychowicz2017hindsight}
M.~Andrychowicz, F.~Wolski, A.~Ray, J.~Schneider, R.~Fong, P.~Welinder,
  B.~McGrew, J.~Tobin, O.~P. Abbeel, and W.~Zaremba, ``Hindsight experience
  replay,'' in \emph{Advances in neural information processing systems}, 2017,
  pp. 5048--5058.

\bibitem{frechet_computing}
T.~Eiter and H.~Mannila, ``Computing discrete fr{\'e}chet distance,'' Citeseer,
  Tech. Rep., 1994.

\bibitem{slerp}
K.~Shoemake, ``Animating rotation with quaternion curves,'' in
  \emph{Proceedings of the 12th annual conference on Computer graphics and
  interactive techniques}, 1985, pp. 245--254.

\bibitem{hausdorff}
A.~A. Taha and A.~Hanbury, ``An efficient algorithm for calculating the exact
  hausdorff distance,'' \emph{IEEE transactions on pattern analysis and machine
  intelligence}, vol.~37, no.~11, pp. 2153--2163, 2015.

\end{thebibliography}

\end{document}